\documentclass[sigconf, nonacm]{acmart}

\usepackage{lipsum}
\usepackage{float}
\usepackage{balance}
\usepackage{enumerate}

\usepackage{lscape} 
\usepackage{subcaption}
\usepackage{xcolor}
\usepackage{colortbl}
\usepackage{algorithm}
\usepackage{algorithmic}

\let\oldnl\nl% Store \nl in \oldnl
\newcommand{\nonl}{\renewcommand{\nl}{\let\nl\oldnl}}% Remove line number for one line

\definecolor{bgred}{RGB}{255,210,205}
\definecolor{bgblue}{RGB}{210,220,255}
\definecolor{bgyellow}{RGB}{240,240,190}
\definecolor{bggrey}{RGB}{223,223,225}
\definecolor{purple}{RGB}{180,0,180}

\definecolor{hvygreen}{RGB}{50,200,0}
\definecolor{hvypink}{RGB}{102,0,51}

\newcolumntype{?}{!{\vrule width 2pt}}

\makeatletter
\def\old@comma{,}
\catcode`\,=13
\def,{%
  \ifmmode%
    \old@comma\discretionary{}{}{}%
  \else%
    \old@comma%
  \fi%
}
\makeatother
    \newcommand{\system}{SubStrat}
    \newcommand{\algo}{Gen-DST}

\begin{document}
%\title{Fast Auto-ML Based on Sub-Table Construction Using the Genetic Algorithm Approach}

%\title{\system{}: Faster AutoML with Measure-Preserving Data Subsets}
\title{\system{}: A Subset-Based Strategy for Faster AutoML}

%%
%% The "author" command and its associated commands are used to define the authors and their affiliations.
\author{Teddy Lazebnik}
\email{t.lazebnik@ucl.ac.uk}
\affiliation{%
  \institution{University College London}
  \country{UK}
}

\author{Amit Somech}
\email{somecha@cs.biu.ac.il}
\affiliation{%
  \institution{Bar-Ilan University}
  \country{Israel}
}

\author{Abraham Itzhak Weinberg}
\email{abraham-itzhak.weinberg@biu.ac.il}
\affiliation{%
  \institution{Bar-Ilan University}
 \country{Israel}
}

\begin{abstract}
Automated machine learning (AutoML) frameworks have become important  tools in the data scientists' arsenal, as they dramatically reduce the manual work devoted to the construction of ML pipelines.
Such frameworks intelligently search among millions of possible ML pipelines - typically containing feature engineering, model selection and hyper parameters tuning steps - and finally output an optimal pipeline in terms of predictive accuracy.

However, when the dataset is large, each individual configuration takes longer to execute, therefore the overall AutoML running times become increasingly high.

To this end, we present \system{}, an AutoML optimization strategy that tackles the data size, rather than configuration space. It wraps existing AutoML tools, and instead of executing them directly on the entire dataset, \system{} uses a genetic-based algorithm to find a small yet representative data \textit{subset} which preserves a particular characteristic of the full data. It then employs the AutoML tool on the small subset, and finally, it refines the resulted pipeline by executing a restricted, much shorter, AutoML process on the large dataset. Our experimental results, performed on two popular AutoML frameworks, Auto-Sklearn and TPOT, show that SubStrat reduces their running times by 79\% (on average), with less than 2\% average loss in the accuracy of the resulted ML pipeline.
\end{abstract}

\maketitle

    %%%%%%%%%%%%%%%%%%%%%
    %      INPUTS       %
    %%%%%%%%%%%%%%%%%%%%%
    \section{introduction}

Automated machine learning (AutoML) frameworks~\cite{karmaker2021automl,gijsbers2019open}
are becoming increasingly popular, as they facilitate the time-consuming, difficult task of developing a machine learning model, allowing even non-expert users to build accurate and robust models for their datasets at hand.
To automatically develop a model, AutoML frameworks compare between millions of ML pipeline configurations, and finally output the optimal pipeline, which typically includes data pre-processing, feature engineering, model selection, and hyper-parameters optimization~\cite{he2021automl}.

Clearly, a naive brute-force search scanning all pipeline configurations is often infeasible~\cite{waring2020automated}, therefore different AutoML frameworks employ a variety of optimizations and search heuristics, such as Bayesian optimization~\cite{hutter2011sequential}, meta-learning~\cite{kim2018auto}, reinforcement learning~\cite{heffetz2020deepline}, and genetic algorithms~\cite{olson2016tpot}, in order to reduce the search space and the number of expensive pipeline executions. 

However, when the training data is large -- each pipeline execution takes longer to run, which can add up to hours of search time, even when using state-of-the-art AutoML frameworks~\cite{he2021automl}. While cloud-based AutoML services\footnote{AutoML is a paid service offered by all major cloud service providers, such as Amazon, Google, Microsoft, as well as by dedicated companies such as H2O.ai and DataRobot.} suggest using better hardware (e.g., larger RAM, more GPUs) when working with large datasets -- this results in much higher costs to the user.  

 To this end, we present \system{}, a new strategy for reducing AutoML computation costs, tackling the \textit{data size} rather than the \textit{configuration search space}.  
 In a nutshell, instead of employing an exiting AutoML tool directly on the entire dataset, we first compute a special \textit{data subset} which preserves some characteristics of the original one. We then employ the AutoML tool over the subset (which is significantly faster), and last, we refine the resulted model configuration by executing a limited, shorter AutoML process over the original dataset.
 
 %To our knowledge, this is the first AutoML optimization strategy that 
 %reduces the size of data rather than the configuration space in order to speedup the AutoML process. 
 The main advantage of our system is the compatibility with state-of-the-art existing AutoML tools -- allowing data scientists to continue using their favorite frameworks while significantly reducing computation times. \textbf{Our experiments show that our system, when employed together with Auto-Sklearn~\cite{auto_sklearn} and TPOT~\cite{tpot}, two of the most popular AutoML frameworks, reduced computation times by an average of 79\%, while retaining 98\% of the best model accuracy.}
 
 \vspace{1mm}
 The main contribution of this work is as follows:
 \begin{enumerate}
\item We present a subset-based optimization strategy for AutoML, aimed at reducing AutoML computation costs with a minimal decrease in model performance. 
\item We introduce the general notion of measure-preserving data subsets, and formulate their generation as an optimization problem. We then formulate a \textit{dataset-entropy} measure and provide a fast and effective genetic algorithm that is able to efficiently generate such entropy-preserving subsets. 
 \item We performed an extensive experimental evaluation over 10 datasets from various domains, and compared our results to 10 different baselines -- using both AutoSklearn and TPOT frameworks. We further discuss the effect of the data subset size on the trade-off between computation times and model accuracy. 
 %\item We make our code and datasets publicly available, and publish them as the first benchmark, to our knowledge, for subset-based AutoML optimizations.
 \end{enumerate}

 \subsection{Problem \& Solution Overview}

In a typical AutoML scenario, a data scientist desires to build an ML model for predicting the value of some target feature $y$ in dataset $D$. 
Rather than manually constructing the model, the data scientist employs an AutoML tool $A$ which intelligently scans multitudes of ML pipelines (i.e., feature engineering, model selection, and hyper-parameters optimizations) and outputs a particular configuration which achieves the highest predictive performance\footnote{Note that other AutoML objectives can be used, such as finding the most compact configuration~\cite{he2021automl}, which is easier to deploy in a production environment.}. We denote the application of AutoML tool $A$ over dataset $D$ to predict the target $y$ by $A(D,y)\rightarrow M^\star$, where $M^\star$ is the best configuration that $A$ could find. 

As mentioned above, the larger the dataset, the higher the computational cost of the AutoML,  since each candidate-pipeline takes longer to execute. Let $Time(M^\star)$ be the time it takes  $A$ to generate $M^\star$, with final model accuracy, denoted by $Acc(M^\star)$.

The goal of \system{}, our subset-based optimization strategy, is to utilize a data \textit{subset} in order to reduce AutoML computation times, while retaining the output model performance. Namely, to generate a model configuration $M_{sub}$ s.t. $Time(M_{sub})<<Time(M^\star)$ but $Acc(M_{sub})\approx Acc(M^\star)$.

\begin{figure}[!t]
\vspace{-2mm}
    \centering
    \includegraphics[width=0.4\textwidth]{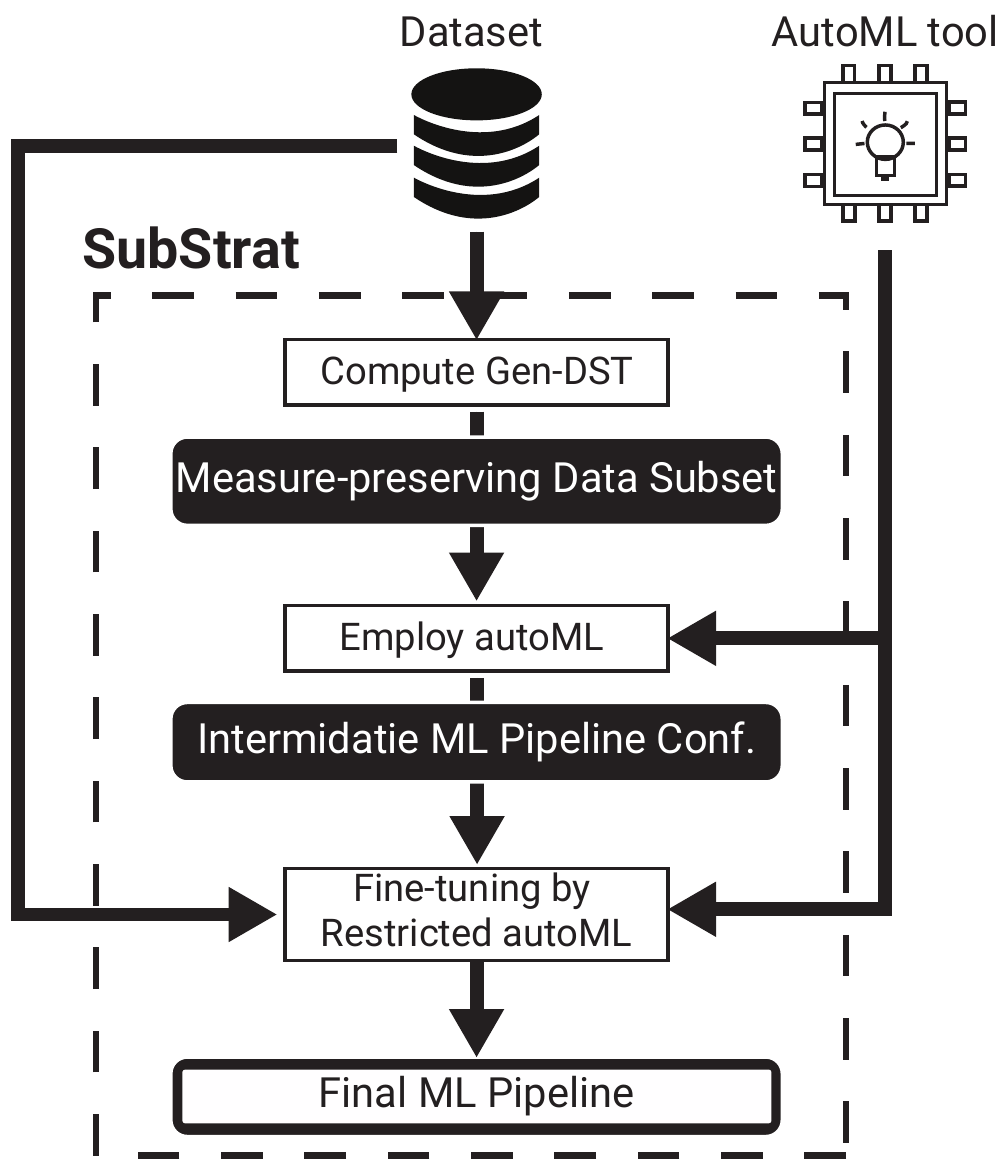}
    \caption{\system{} Workflow}
    \label{fig:flow}
    \vspace{-3mm}
\end{figure}

\vspace{1mm}
Abstractly, given a dataset $D$ of size $N \times M$ and a target feature $y$, \system{} works in three steps (See Figure~\ref{fig:flow} for illustration): 
\begin{enumerate}
    \item Find a small data subset $d$, of size $n\times m$, s.t. $n<<N$ and $m<<M$.
    \item Employ the AutoML tool over $d$, i.e., $A(d,y)\rightarrow M'$.
    \item Fine-tune the intermediate pipeline configuration $M'$, by employing a restricted, faster instance of $A$ back on $D$ to obtain the final configuration $M_{sub}$.
\end{enumerate}

Although it is quite obvious that employing AutoML on a fraction of the data takes less time, finding an adequate subset in a timely fashion is challenging.

For instance, one could easily take a random subset of the data, and employ AutoML over it. Unfortunately, as further discussed in Section~\ref{sec:exp_baseline_results}, using such random subsets in our framework reduces the final model accuracy by more than 27\% compared to the accuracy of $M^\star$. 

While 27\% accuracy loss in ML is unanimously considered too low, there is an ongoing discussion about the acceptability of model accuracy for different applications in light of other objectives such as interpretability, and training time (See, e.g.,~\cite{kay2015good,weinberg2019selecting,gupta2016model}). Following these discussions, in this work we assume that \textit{a decrease of more than 5\% in accuracy is largely unacceptable for AutoML}.

\paragraph*{Solution \& Paper Outline} 
We begin by reviewing related work (Section~\ref{sec:related}).
We then describe the architecture and methods of \system{} in Section~\ref{sec:solution}: we first introduce the notion of measure-preserving data subsets, which are designed to capture qualities of the original data (Section~\ref{sec:dts}). Next, since finding the optimal measure-preserving subset is computationally infeasible, we formulate an optimization problem (Section~\ref{sec:optimization}), and present a genetic-based algorithm to efficiently solve it (Section~\ref{sec:genetic}). Last, we discuss our method for fine-tuning the intermediate model configuration, adapting it to fit the large, original dataset (Section~\ref{sec:finetune}). 
Our experimental evaluation is brought in Section~\ref{sec:experiments}, and we conclude in Section~\ref{sec:conclusion}.

\section{Related Work}
\label{sec:related}

We survey related works in the field of AutoML as well as other works which aim to reduce datasets' size in different contexts. 

\paragraph*{Automated Machine Learning (AutoML)}

As mentioned above, Auto-ML is a growing research field~\cite{he2021automl}, aiming to automate the process of developing ML models for a given task and dataset~\cite{real2020automl}.
Since ML pipelines typically contain several steps (such as feature engineering, model selection, and hyper-parameters tuning), the space of possible configuration is too large for a naive search. Therefore, existing AutoML can be roughly divided into two main categories: search-space optimizations and meta-learning solutions. 

Search space optimizations employ intelligent search strategies and heuristics in order to perform the configuration selection more efficiently on ad-hoc datasets. Example methods used are Bayesian optimization~\cite{feurer2020auto,hutter2011sequential}, directed search~\cite{wu2021frugal,wang2021flaml} and genetic programming~\cite{tpot}. 
Meta-learning solutions for AutoML~\cite{drori2021alphad3m,heffetz2020deepline} take a different approach in order to produce an optimal pipeline configuration -- by training, in advance, an ML model on a large corpus of datasets, then predicting the optimal configuration given the dataset and task at hand. This solution, while significantly faster, is more resource-intensive and assumes the user has a suitable collection of datasets to train on~\cite{wang2021flaml}. 
In particular, we note the works in~\cite{feurer2015efficient,feurer2020auto}, describing the popular Auto-Sklearn system, which combines the two approaches and uses meta-learning alongside search optimizations to obtain further speedup.

Differently from these works, \textit{the goal of \system{} is to reduce the size of the input dataset, rather than the space of pipeline-configurations.}
It is designed to improve the running-time of existing search-based AutoML tools by running the majority of computation on a significantly smaller data subset, discovered by our genetic-based algorithm (See Section~\ref{sec:genetic}). 

\paragraph*{Data Reduction Techniques}
Reducing the dataset size is considered in previous work, where numerous methods are suggested for selecting either rows or columns (features). 

Feature selection~\cite{chandrashekar2014survey,info_entropy_feature_ranking} is a prominent step in many ML pipelines, where the goal is to reduce the number of input variables considered by the model. This is done in order to reduce training times as well as the complexity of the model. There is a plethora of research works (See~\cite{chandrashekar2014survey} for a survey), roughly categorized as Filter-based techniques, that yield the Top-k features in terms of a given metric (e.g., Chi-Square, ANOVA and Information-Gain)~\cite{info_entropy_feature_ranking}; as well as Embedded and Wrapper methods, which directly utilize the ML models to determine the important features~\cite{cichocki2014era}.
Selecting dataset rows is also widely considered in previous research, for either general-purpose methods that produce a norm-preserving sub-matrix~\cite{cohen2015lp} or for specific tasks such as search-results diversification~\cite{drosou2010search} and faster generation of data visualizations~\cite{vizsample}. The latter use dedicated, task-dependent utility definitions. 

\system{} is different from these works as it generates data subsets by selecting both rows \textit{and} columns, hence solving a different, more complex optimization problem. We show in our experimental evaluation, that data subsets composed by separately applying a feature-selection and row-sampling methods yield inferior results to the ones generated by \system{}.

\section{Solution Architecture}

\label{sec:solution}
We next describe the components of \system{} in more detail.

\subsection{Measure-Preserving Data Subsets}

\label{sec:dts}
As mentioned above, our goal is to find a subset of the original dataset which preserves a particular characteristic of the data.

Let $D$ be a dataset of $N$ rows and $M$ columns.
Denote its row and column indices by $R=1,2,\dots,N$ and $C=1,2,\dots,M$, respectively. Intuitively, a data subset (referred to as DST, for short) of a full dataset $D$ is simply a subset of the rows of $D$, projected over a subset of the columns.

\begin{definition}[Data Subset (DST)]
Given a dataset $D$ with row-indices $R$ and column-indices $C$,
a data subset of size $n \times m$ is defined as follows. 
Let $[R]^n$ be the set of all $n$-subsets of $R$, i.e.,  $[R]^n = \{R'|R' \subseteq R ~\wedge~| R |=n\}$, and $[C]^m$ be the set of all $m$-subsets of $C$.
Then, given  $r\in [R]^n$ and $c\in [C]^m$, the data subset is defined by
$D[r,c]$, i.e., the rows in $D$ indicated in $r$, projected over the columns indicated in $c$.
For convenience, we simply denote a DST by $d$ when possible. 
\end{definition}

Last, since the target column is crucial for the AutoML process,
our framework automatically inserts it into every DST. We restrict our framework to only consider data subsets which contain the target column. 

\begin{example}
Consider the 10X5 dataset in Table~\ref{table:example}, taken from the \textit{flight service review} dataset in our experiments (See Section~\ref{sec:exp_setup}). The green and red cells represent two different 5X3 data subsets: $d_{green}=D[(1,2,3,6,8)(1,4,5)]$ and $d_{red}=D[(4,5,7,9,10),(2,3,5)]$.
Note that both contain the target column (the right-most column in Table~\ref{table:example}). 
\end{example}

\begin{table}[!t]
\centering
\begin{tabular}{|l|l|l|l|l|l|}
\hline
 & \textbf{Age} & \textbf{Gender} & \begin{tabular}[c]{@{}l@{}}\textbf{Flight}\\ \textbf{distance}\end{tabular} & \begin{tabular}[c]{@{}l@{}}\textbf{Delay}\\ \textbf{[minutes]}\end{tabular} & \begin{tabular}[c]{@{}l@{}}\textbf{Satisfied}\\ \textbf{(target)}\end{tabular} \\ \hline
\(\boldsymbol{R_1}\) & \cellcolor{green}25 &  1 &  460 & \cellcolor{green}18 & \cellcolor{green}1 \\ \hline
\(\boldsymbol{R_2}\) &  \cellcolor{green}62 &  1 &  460 &  \cellcolor{green}0 & \cellcolor{red}\cellcolor{green}0 \\ \hline
\(\boldsymbol{R_3}\) &  \cellcolor{green}25 &  0 &  460 &  \cellcolor{green}40 & \cellcolor{green}1 \\ \hline
\(\boldsymbol{R_4}\) & 41 & \cellcolor{red}0 & \cellcolor{red}460 & 0 & \cellcolor{red}1 \\ \hline
\(\boldsymbol{R_5}\) & 27 & \cellcolor{red}1 & \cellcolor{red}460 & 0 & \cellcolor{red}1 \\ \hline
\(\boldsymbol{R_6}\) & \cellcolor{green}41 & 1 & 1061 & \cellcolor{green}0 & \cellcolor{green}0 \\ \hline
\(\boldsymbol{R_7}\) & 20 & \cellcolor{red}0 & \cellcolor{red}1061 & 0 & \cellcolor{red}0 \\ \hline
\(\boldsymbol{R_8}\) & \cellcolor{green}25 & 0 & 1061 & \cellcolor{green}51 & \cellcolor{green}0 \\ \hline
\(\boldsymbol{R_9}\) & 13 & \cellcolor{red}0 & \cellcolor{red}1061 & 0 &  \cellcolor{red}1 \\ \hline
\(\boldsymbol{R_{10}}\) & 52 & \cellcolor{red}1 & \cellcolor{red}1061 & 0 & \cellcolor{red}1 \\ \hline
\end{tabular}
\caption{An example dataset with two 5X3 subsets marked in green and red. $d_{green}$ is a measure-preserving subset (w.r.t. the dataset-entropy measure), while $d_{red}$ is not.}
\label{table:example}
\vspace{-3mm}
\end{table}

\vspace{1mm}
As will be shown in Section~\ref{sec:experiments}, simply using a random DST (in which the row and column subsets are chosen uniformly at random) in our solution induces a substantial decrease in the accuracy of the AutoML process. 
Our goal is therefore to find a more \textit{representative} DST, that preserves some characteristic of the original dataset.
Let $F:\mathbb{D}\rightarrow \mathbb{R}$ be a \textit{dataset-measure}
which takes a dataset as input and evaluates a characteristic of it by a real number. 

We define a measure-preserving DST as follows.

\begin{definition}[Measure-Preserving DST]
Given a dataset $D$, a DST $d=D[r,c]$, and a dataset-measure $F:\mathbb{D}\rightarrow \mathbb{R}$, we call a DST $d$ \textit{measure-preserving} if
$F(d)\approx F(D)$.
\end{definition}

While any measure that evaluates a characteristic of the data may be applicable, in this work we use a \textit{dataset entropy} function, which assesses the ``amount of information'' conveyed in the data. Entropy-based functions are widely used to characterize data, for tasks such as feature selection and decision-tree splits~\cite{info_entropy_feature_ranking}. In our context, we define dataset entropy as follows.

\begin{definition}[Dataset Entropy]
Given dataset $D$ of size $N\times M$, Let $D_{ij}$ be the value in row $i$ and column $j$.
$$H(D) = \frac{\sum_{j=1}^{M} \left(\-\sum_{i=1}^{N}  P_j(D_{ij})\cdot Log_2P(D_{ij}) \right)}{M} $$
Where $P_j(D_{ij})$ is a probability function corresponding to the frequency of the value in $D_{ij}$ w.r.t. Column $j$. Namely, if $D_{ij}=v$ then: $$P_j(v)=\frac{\sum_{k=1}^{N} I[D_{kj}=v]} {N}$$
\end{definition}

\vspace{1mm}
\begin{example}
Consider again the dataset and two subsets depicted in Table~\ref{table:example}. 
Calculating the dataset entropy we obtain: $$H(D)=\frac{2.65+1+1+1.4+0.97}{5} =1.395$$
We indeed observe that $D$ contains two columns with high entropy (`Age' and `Delay'). 
These columns are also selected in the green DST, which obtains the score: $$H(d_{green}) = \frac{1.37+1.92+0.97}{3} = 1.42$$
$d_{green}$ is the 5X3 DST which obtains the closest dataset-entropy score to $D$.
However, the red DST, which contains low-entropy columns, obtains a lower score of $H(d_{red}) =0.89$. 
Hence, $d_{green}$ is considered a measure-preserving DST, whereas $d_{red}$ is not.
\end{example}

\vspace{1mm}
Last, note that while dataset-entropy worked best in our experiments, our optimization algorithm, as described below, is generic and can take other possible dataset measures as input (e.g., $p$-norm, mean-correlation, and coefficient of variation). 

\subsection{DST as an Optimization Problem}

\label{sec:optimization}
Ideally, we would like to find the best-preserving DST for a dataset $D$.
Namely, the best DST of size $n \times m$ can be found by:
\[ \underset{r\in [R]^n, c\in [C]^m }{\operatorname{argmin}} \left| F\left(D[r,c]\right) - F(D) \right| \]

If $n$ and $m$ are small, then finding the best-preserving DST can be done in 
$O(N^n\cdot M^m)$ time, by a brute-force search that traverses through all possible DST of size $n \times m$. Clearly, this becomes infeasible for large datasets or when a larger DST is needed (we found that a DST of size \((\sqrt{N}, 0.25M)\) yields a good balance between accuracy-loss and running-times).

We therefore define an optimization problem, which is to minimize the difference between the DST and the original dataset,
i.e., 
$$\mathcal{L}(r,c) = \left| F\left(D[r,c]\right) - F(D) \right|  $$

Note that while numerous methods and algorithms can be used to minimize $\mathcal{L}(r,c)$ (See Section~\ref{sec:exp_baseline_desc}), we must use an approach that also obtains short convergence times. Otherwise, the optimization process will take too long, hence diminishing the efficacy of our overall solution in reducing AutoML running times.   

\subsection{A Genetic-Based Algorithm for Finding DST}

\label{sec:genetic}
Our framework employs a Genetic Algorithm (GA), a well-known and commonly-used meta-heuristic search method, based on the biological theory of evolution \cite{ga_intro}. Briefly, GA simulates \textit{evolution} through a natural selection process: First, a population of \(\phi\) candidate-solutions, each comprising a set of properties, referred to as \textit{genes} is selected at random. The algorithm then iteratively mutates and alters the genes in order to create ``better'' solutions w.r.t. a \textit{fitness function}, which corresponds to the optimization objective. In particular, at each \textit{generation} (i.e., iteration), the GA typically performs several stochastic operators~\cite{cross_over,selection_operator}: (1) a \textit{mutation} operator which induces random noise into the genes of a candidate-solution, (2) a \textit{cross-over} operator which combines the genes of two candidate-solutions, and (3) a \textit{selection} operator which refines the population of the next generation, by keeping fitter candidate-solutions with higher probability than less-fitting solutions. Finally, after a number of generations (\(\psi\) - chosen according to a predefined stopping criteria), the fittest candidate-solution is selected as the output of the GA algorithm.

We next describe ~\algo{}, our genetic-based algorithm for finding measure-preserving DSTs, as depicted in Algorithm~1.
\begin{algorithm}[t]
\label{algo:gen}
	\caption{\textbf{\algo{}}} \label{alg0}
	\begin{algorithmic}[1]
	    \STATE $\text{\textbf{Hyper-parameters: }} (\psi), (\phi), (\xi), (\gamma)$
	    \STATE $\text{\textbf{Input: }} \text{dataset }  (D) , \text{dataset-measure } (F), \text{DST size } (n,m)$
	    \STATE $\text{\textbf{Output: }} \text{data subset }  (d)$
	    \STATE $P_0 \Leftarrow \text{generate } (\phi) \text{ candidates in random }$
	    \STATE $best\_dst \Leftarrow argmax_{G \in P_0} F(G, D)$
	    \FOR{$\text{generation}~i \in [1, \dots, \psi]$}
  	        \STATE  $P_i \Leftarrow Mutation\_Operator(P_i, \xi, p_{rc})$
  	        \STATE  $P_i \Leftarrow Crossover\_Operator(P_i, p_m)$
  	        \STATE  $P_{i+1} \Leftarrow Selection\_Operator(P_i, \alpha)$
  	        \IF{$\max_{G \in P_{i+1}} F(G, D) > F(best\_dst, D)$}
  	            \STATE $best\_dst \Leftarrow argmax_{G \in P_{i+1}} F(G, D)$
  	        \ENDIF
  	    \ENDFOR
  \STATE return $d := D[best\_dst(r), \; best\_dst(c)]$
	\end{algorithmic}
\end{algorithm}

\vspace{1mm}
\noindent\textbf{Genetic representation of candidate-DSTs.} 
The genetic representation of a candidate-DST, denoted $G$, comprises of $n+m$ chromosomes: $n$ row-chromosomes, that correspond to $n$ row indices of dataset $D$, and $m$ column-chromosomes, that correspond to $m$ column-indices.
More formally,
$$G\coloneqq (r,c), ~ r \in  [R]^n, c \in [C]^m $$ 
Recall that for a dataset $D$,  $R$ and $C$ denote its row and column indices.

\vspace{1mm}
\noindent\textbf{Fitness Function.} The fitness function $f(G)$ is simply the negative loss of the DST-candidate $G=(r,c)$, Namely,
$$f(G) \coloneqq -\mathcal{L}(r,c) = -\left| F\left(D[r,c]\right) - F(D) \right|  $$

\vspace{1mm}
\noindent\textbf{\algo{}  Workflow \& Operators.}
\algo{}, as depicted in Algorithm~1, works as follows. First, an initial population \(P\) of candidate DSTs is randomly generated, s.t. each candidate-DST $G(r,c)$ contains the target column $y$, i.e. $t\subset c$.
Then, for each generation $i$, we perform (1) mutation, (2) cross-over and (3) selection, in order to generate the next-generation population $P_{i+1}$:

\vspace{1mm}
\noindent\textit{(1) Mutation.} The mutation operator is stochastically employed, for each candidate-solution $G=(r,c)$ in the population $P_i$ with probability $\xi$.
    First, we randomly decide if to mutate rows or columns w.r.t. probability $p_{rc}$. If, for example, a row-mutation is decided upon, we randomly replace one of the row-indices in $r$. Namely, we mutate $G$ and form $G'$ s.t. $$G(r',c),~r' \in [R]^n ~\wedge~ |r \cap r'| = n-1 $$ 
    A similar process is performed for column mutations, only that the target column $y$ cannot be mutated.  
    
\vspace{1mm}

\noindent\textit{(2) Cross-Over.} Cross-over is employed for two candidate-DSTs $G_a=(r_a,c_a)$ and $G_b=(r_b,c_b)$ in population $P_i$ with the goal of creating two next-generation DSTs, $G_{ab}$ and $G_{ba}$. We begin by selecting whether to cross rows or columns (similar to the mutation operator), with probability $p_{rc}$. 
    Then, assuming (w.l.o.g.) that columns cross-ever is selected, we randomly choose a split-size $1< s < m$, and use it to split both $c_a$ and $c_b$, each to two random subsets - one of size $s$ and one of size $m-s$, i.e., $c_a = c_a^s \cup c_a^{m-s}$ and $c_b = c_b^s \cup c_b^{m-s}$. 
    The cross-over then unifies complementing subsets from $a$ and $b$, creating $c_{ab}$ and $c_{ba}$:
    $$c_{ab}=c_a^s \cup c_b^{m-s},~ c_{ba} = c_b^s \cup c_a^{m-s}$$
    Finally, the next-generation DSTs are set as $G_{ab} = (r,c_{ab})$ and $G_{ba} = (r,c_{ba})$\footnote{In case the size of  $c_{ab}$ or $c_{ba}$ is smaller than $m$, we insert the required amount of columns at random, while also making sure the target column $y$ is contained in both.}.
    
    The cross-over operation is  performed over the entire population $P_i$: $P_i$ is first split to disjointed pairs of candidate-DSTs, then the cross-over is performed on each such pair.

\vspace{1mm}
\noindent\textit{(3) Selection.} Last, after employing mutation and cross-over, we employ the selection operator which forms the next-generation population $P_{i+1}$. We use the \textit{royalty tournament} 
    operator~\cite{selection_operator}, which selects the best $\alpha \cdot \phi$ candidate-DSTs from $P_i$ according to the fitness function $f(G)$. The rest of the $\phi(1-\alpha )$ DSTs are sampled (with repetitions) according to their fitness score, i.e., with probability:
    $$ p_{select}(G) = \frac{f(G)}{\sum_{G' \in P_i} f(G')}   $$.
    
Last, the stopping criterion of \algo{} is either reaching a predefined limit on the generations number, or a convergence criterion which stops the execution when the fittest DST of population $P_{i+1}$ is not significantly better than the fittest solution in $P_{i}$. In this case, we return the DST that obtained the highest fitness score, over all previous generations.

\subsection{Fine-Tuning the Intermediate Configuration}
\label{sec:finetune}
Recall that after generating a DST $d$, using our algorithm described above, we employ the AutoML tool on $d$ to obtain an intermediate pipeline configuration, i.e. $A(d,y)\rightarrow M'$.
This configuration needs to be adapted back to fit dataset $D$. 

To do so, we employ $A$ back on $D$, but restrict its search space, forcing it to only consider configurations that use the same ML model as specified in $M'$. 
Such a simple yet efficient restriction allows \system{} to retain short running times, while  considerably improving the intermediate configuration $M'$, reaching almost the same accuracy of the best configuration $M^\star$.

\section{Experiments}
\label{sec:experiments}
We conducted a thorough experimental study with the goal of examining the effectiveness of \system{} in reducing the running times of existing AutoML tools while retaining the accuracy of their output ML pipelines. 

\subsection{Setup \& Methodology}
\label{sec:exp_setup}

\paragraph*{Experimental Framework \& Methodology}
Given an input dataset and a target feature, we first directly employ an AutoML tool and obtain its output ML pipeline configuration.
Recall that the AutoML tools (we use the popular AutoSklearn and TPOT frameworks, as described below) apply sophisticated algorithms to prune non-promising ML pipeline configurations, and finally output the pipeline with the highest predictive accuracy on the specified target feature. 
We record both the running time and the accuracy of the resulted model, which serve as our primary baseline, denoted Full-AutoML.

We then examine whether our subset based strategy can indeed reduce AutoML running times, and still generate ML pipelines as accurate as Full-AutoML. 
To generate the data subsets, we used \algo{} as well as 10 other baselines (see below). 
For each instance, we compute the relative running time and accuracy w.r.t Full-AutoML. We report the following metrics: \textit{time-reduction}, which indicates how much time was saved:
$$Time\text{-}Reduction = 1 - \frac{Time(M_{sub})}{Time(M^\star)}$$  
 We also report the \textit{relative-accuracy},
 indicating the proportion of accuracy of Full-AutoML that was successfully retained:  
 $$Relative\text{-}Accuracy = \frac{Acc(M_{sub})}{Acc(M^\star)}$$
 
 Recall that following the discussions in ~\cite{kay2015good,weinberg2019selecting,gupta2016model}), 
 we assume that a relative accuracy of less than 95\% is largely unacceptable. 
 
\paragraph*{Datasets}
We used 10 popular datasets from Kaggle~\cite{kaggle} and UCI Machine Learning Repository~\cite{uci}. The datasets, as depicted in Table~\ref{table:datasets}, come from various data domains, and have different shapes (up to 1M rows / 123 columns). Links to the full datasets can be found in our code repository~\cite{ourgit}.

\paragraph*{Auto-ML methods}
We evaluated \system{} using both Auto-Sklearn and TPOT, two highly-popular AutoML tools. The two have substantially different underlying technology: \textbf{(1) Auto-Sklearn\cite {feurer2015efficient,feurer2020auto}}, an industry standard tool which uses Bayesian optimization methods together with meta-learning. It works on top of the Python Scikit-Learn library~\cite{scikit-learn}, and generates an ML pipeline configuration comprising of feature prepossessing, ML model selection, and hyper parameters optimization. \textbf{(2) Tree-Based Pipeline Optimization Tool (TPOT) \cite{olson2016evaluation}}, which also utilizes Scikit-Learn as Auto-Sklearn, but uses a genetic programming approach to explore the configuration search-space.   

\paragraph*{Implementation \& Hardware}
\system{} and the rest of the baseline algorithms were implemented in Python 3. Our source code is fully available in~\cite{ourgit}.
We ran the experiments on an Ubuntu Server with an Intel Core i7-9700K CPU and a 64GB RAM. All experiments were conducted in a linear fashion and no other programs were executed on the device except for the operation system. 

\begin{table}[!t]
{\small
\centering
\begin{tabular}{|c|c|c|c|}
\hline
\textbf{Symbol} & \textbf{Domain} & \textbf{\#Rows} & \textbf{\#Columns} \\ \hline
$D_{1}$ & Flight service review & 129880 & 23 \\ \hline
$D_{2}$ & Signal processing & 15300 & 5 \\ \hline
$D_{3}$ & Car insurance & 10000 & 18 \\ \hline
$D_{4}$ & Mushroom classification & 8125 & 23 \\ \hline
$D_{5}$ & Air quality & 57660 & 7 \\ \hline
$D_{6}$ & Bike demand & 17415 & 9 \\ \hline
$D_{7}$ & Lead generation form & 7000 & 15 \\ \hline
$D_{8}$ & Myocardial infarction & 1700 & 123 \\ \hline
$D_{9}$ & Heart disease & 79540 & 7 \\ \hline
$D_{10}$ & Poker matches & 1000000 & 15 \\ \hline
\end{tabular}
\caption{Dataset descriptions and properties}
\vspace{-5mm}
\label{table:datasets}
}
\end{table}

\subsection{Baseline Methods}
\label{sec:exp_baseline_desc}

We implemented 10 different baselines in 6 different categories (A-F), as depicted in Table~\ref{table:algorithms}.
To clarify the scope of comparison, recall again that AutoML methods require the raw data as input, therefore any approach that \textit{alters} the data (e.g., PCA, embedding) is inapplicable. Also, we only compare \system{} to other methods for reducing the data size rather than the configuration space, as the latter is performed by the chosen AutoML tool. 
The baselines in categories A-F were therefore devised to answer the following questions:
\begin{enumerate}[i.]
\item Can a trivial, random DST perform well enough? (Category A)
\item Can we use different, existing optimizations for finding measure-preserving DSTs? (Categories A-C)
\item Can we generate effective DSTs using existing techniques for row sampling and column selection? (Categories D-E)
\item Can \system{} obtain good performance without the fine tuning phase? (Category F) 
\end{enumerate}

\vspace{1mm}
\noindent\textit{A. Monte-Carlo Search.}
We began with a simple random search technique, which given a predefined time/iteration budget $B$, randomly generates DSTs, calculates their measure-preserving loss (as defined in Section~\ref{sec:optimization}), and at the end of the time limit (or max iteration) returns the DST that obtained the minimal loss. We use three instances with different budgets: (1) \textit{MC-100}, which examines 100 DSTs, (2) \textit{MC-100K}, designed to have approximately the same running-times as \algo{}, allowing it to compare about 100K DSTs. Last, to demonstrate the optimization challenge of finding DSTs, we also examined (3) \textit{MC-24H}, which stops after 24 hours. While the latter cannot improve the running time of AutoML, we examine its performance only in terms of relative accuracy.  

\vspace{1mm}
\noindent\textit{B. Multi-Arm Bandit.}
Additionally, we implemented a more sophisticated \textit{Multi-Arm Bandit (MAB)} baseline, which also attempts to find a measure-preserving DST. MAB is a well-known search framework which balances exploration and exploitation within the search space\cite{li2005multi}. 
We implemented the MAB baseline by formulating two types of arms: row-arms and column-arms.
At each round, the model needs to choose  $n$ rows and $m$ columns, and balance the exploration/exploitation of its choices using an $\epsilon$-greedy policy.  

\vspace{1mm}
\noindent\textit{C. Greedy Selection.}
Another possible optimization is to use a greedy selection process. Since the loss is dependent on both the rows and the columns, we used two instances of the algorithm:
(1) \textit(Greedy-Seq) which first selects $n$ rows and then $m$ columns. 
The $n$ rows are found in a greedy manner, s.t. at each step we add to the DST $d$ a new row from $D$ which locally diminishes the local loss of $d$ (while using all columns in $D$).
In the second step we choose $m$ columns in a similar manner, only that the loss is computed w.r.t. the rows already found in the row-selection phase. 
We also implemented (2) \textit{Greedy-Mult} which attempts to greedily select both a row and a column at each step. 

\vspace{1mm}
\noindent\textit{D. Clustering-Based Approach.} 
This method does not attempt to find measure-preserving DSTs, yet tries to select representative rows and columns using clustering. 
The \textit{KM Baseline} first clusters the rows in $D$ to $n$ clusters, by employing K-means clustering~\cite{likas2003global}. Then, to choose $n$ representative rows, we pick the ones that are the closest to each of the $n$ cluster centroids.
To select $m$ columns, we do the same process by applying K-Means on the column vectors. 

\vspace{1mm}
\noindent\textit{E. Information-Gain (Feature Selection).}
Information-gain (IG) is a commonly used technique for feature selection \cite{kraskov2004estimating}. Similarly to our dataset-entropy measure, it is also based on entropy calculations, where the goal is to select $m$ columns that have the highest IG, w.r.t. the target feature $y$. Intuitively, these are the columns that provide the most ``information'' about $y$. As IG can only be used for feature-selection, we implemented two different baselines here: (1) \textit{IG-Rand} which selects columns using $IG$ and chooses the rows at random, and (2) \textit{IG-KM}, which uses IG for column selection, and the KM baseline to choose the rows.

\vspace{1mm}
\noindent\textit{F. \system{} Without Fine-Tune.}
Last, we examine the importance of the fine-tuning phase, by using a limited version of \system{}, denoted \system{}-NF. This version outputs the intermediate configuration $M'$ -- resulted by applying the AutoML tool only on the DST generated by \algo{}, without employing fine tuning on the full dataset.

\vspace{2mm}
\noindent\textit{Baselines Default Configurations.}
All hyper parameters for \system{}, as well as the baselines, were set using a grid search, optimizing on the harmonic mean of time-reduction and relative-accuracy. The configurations also included the DST size (with varying lengths and widths).
For \system{} the default configuration is: $\psi =30,\phi=100,\xi=0.025,\alpha=0.05,p_{rc}=0.9$, with a 
DST Size of $(\sqrt{N}, 0.25M)$.
See our source code repository~\cite{ourgit} for the baselines' configurations. We further discuss the effect of the DST size on performance in Section~\ref{sec:exp_dst_size}, and alternative configurations in Section~\ref{sec:exp_tradeoff}.

\begin{table}[!t]
\centering
{\small
\begin{tabular}{|l|l|l|}
\hline
\textbf{Cat.} &\textbf{Baseline} & \textbf{Symbol} \\ \hline
%\system{} Algorithm & Gen-DST   \\ \hline
A & Monte-Carlo Search with \(100\) samples & MC-\(100\)  \\ \hline
A & Monte-Carlo Search with \(10^5\) samples & MC-\(100K\) \\ \hline
A & Monte-Carlo Search with 24 hours budget & MC-\(24H\)  \\ \hline
B & Multi-Arm Bandit & MAB \\ \hline
C & Greedy Selection (sequence )& Greedy-Seq \\ \hline
C & Greedy Selection (row+columns) & Greedy-Mult \\ \hline
D & K-Means Clustering & KM  \\ \hline
E & Information Gain (Columns) + random (rows) & IG-Rand  \\ \hline
D, E & Information gain (Columns) + clustering (rows) & IG-KM  \\ \hline
F & \system{} Without Fine Tune & \system{}-NF  \\ \hline
\end{tabular}
\caption{Baseline Methods for Generating Data Subsets} 
\vspace{-4mm}
\label{table:algorithms}
}
\end{table}

 \begin{figure*}[t]
 \vspace{-4mm}
      \begin{minipage}[t]{0.35\textwidth}
   % \begin{figure}[!t]
        \centering
        \includegraphics[width=1\textwidth]{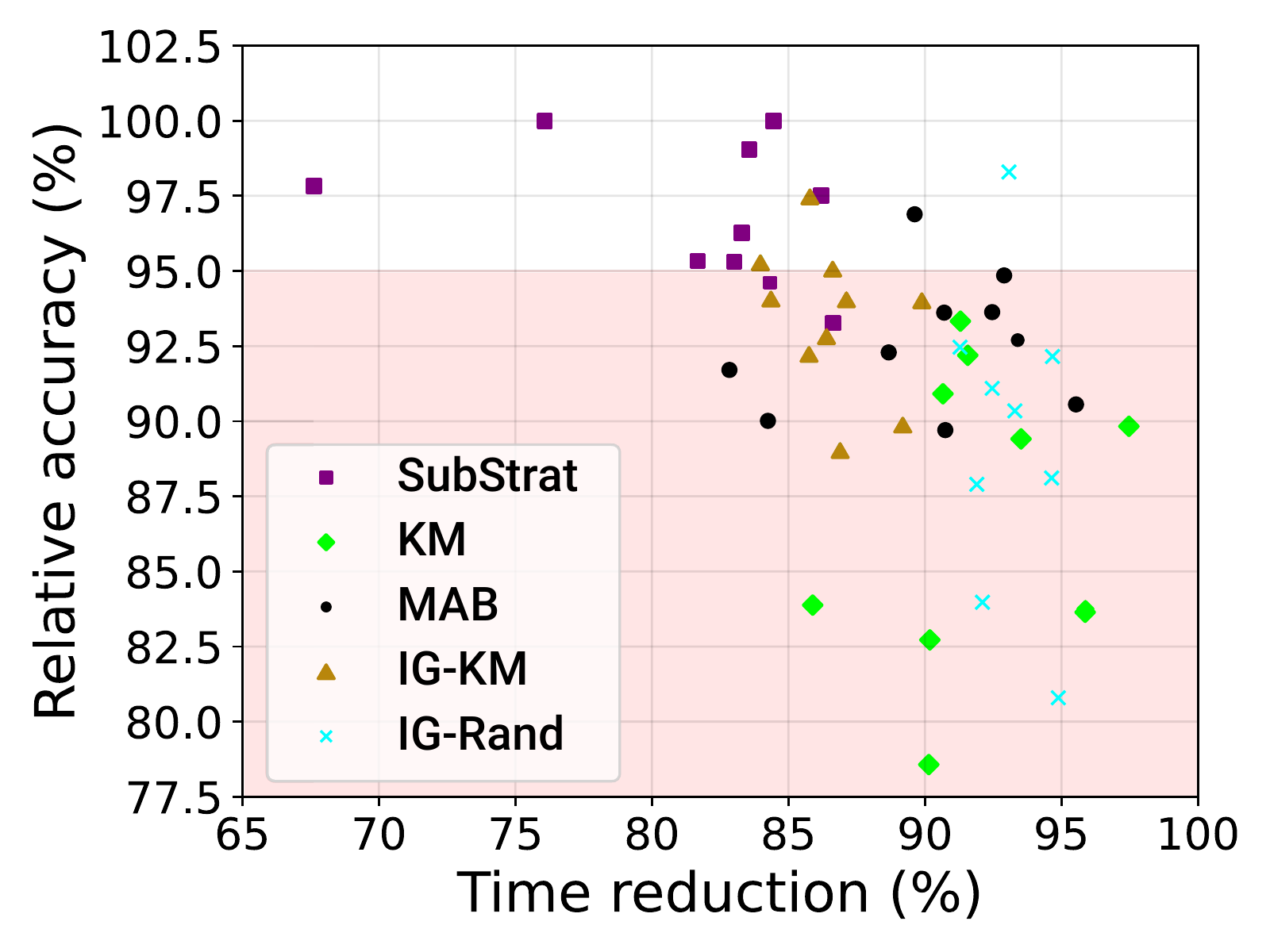}
        \vspace{-9mm}
        \captionof{figure}{Per-Dataset Performance}
        \label{fig:scatter}
    %\end{figure}
    \end{minipage}
    ~~~~
\begin{minipage}[t]{0.64\textwidth}
    %\begin{table*}[!t]
\centering
\vspace{-44.5mm}
\begin{tabular}{|l|ll|ll|}
\hline
\textbf{Algorithm} & \textbf{AutoSklearn} & \textbf{} & \textbf{TPOT} & \textbf{} \\ \cline{2-5} 
\textbf{} & \multicolumn{1}{l|}{\textbf{Time Reduction}} & \textbf{Rel. Acc.} & \multicolumn{1}{l|}{\textbf{Time Reduction}} & \textbf{Rel. Acc.} \\ \hline
\system{} & \multicolumn{1}{l|}{\(81.10 \pm 1.27\)\% } & \(98.71 \pm 0.64 \)\% & \multicolumn{1}{l|}{\(76.57 \pm 1.42 \% \)} & \(97.17 \pm 0.59\)\%  \\ \hline
IG-KM & \multicolumn{1}{l|}{\(86.63 \pm 1.04 \%\) }  & \(93.83 \pm 0.88\)\%  & \multicolumn{1}{l|}{\(85.80 \pm 0.86\)\% } & \(94.42 \pm 0.47\)\% \\ \hline
MAB & \multicolumn{1}{l|}{\(90.95 \pm 1.20\)\% }  & \(92.88 \pm 1.01\)\% & \multicolumn{1}{l|}{\(90.71 \pm 1.33\)\% } & \(92.66 \pm 1.19\)\%\\ \hline
\system{}-NF& \multicolumn{1}{l|}{\(90.99 \pm 0.62\)\% } & \(93.09 \pm 1.12 \)\% & \multicolumn{1}{l|}{\(88.35 \pm 1.42 \% \)} & \(92.08 \pm 1.29\)\%  \\ \hline
IG-Rand & \multicolumn{1}{l|}{\(93.10 \pm 0.49\)\% } & \(89.04 \pm 1.59\)\% & \multicolumn{1}{l|}{\(92.03 \pm 0.48\)\% } & \(87.27 \pm 1.98\)\% \\ \hline
KM & \multicolumn{1}{l|}{\(92.25 \pm 0.97\)\% }  &  \(86.82 \pm 0.91\)\% & \multicolumn{1}{l|}{\(91.32 \pm 0.75\)\%} &  \(87.23 \pm 0.51\)\% \\ \hline
MC-100K & \multicolumn{1}{l|}{\(83.46 \pm 0.30\)\% } & \(78.14 \pm 3.19\)\% & \multicolumn{1}{l|}{\(79.20 \pm 0.34\)\% } &  \(76.65 \pm 3.57\)\% \\ \hline
MC-100 & \multicolumn{1}{l|}{\(97.03 \pm 0.03\)\% } & \(69.57 \pm 4.02\)\% & \multicolumn{1}{l|}{\(96.86 \pm 0.04\)\% } & \(70.08 \pm 3.68\)\% \\ \hline
\end{tabular}
%\end{table*}
\vspace{2mm}
\captionof{table}{Mean Time-Reduction and Relative-Accuracy (Rel. Acc.) Scores}
\label{tab:baselines_results}
    \end{minipage}
    \vspace{-3mm}
\end{figure*}

\subsection{Overall Baseline Comparison Results}
\label{sec:exp_baseline_results}
 Table~\ref{tab:baselines_results} depicts the time reduction and relative accuracy scores obtain by each baseline approach (for both, higher is better).
Each baseline was executed 5 times on each dataset, and the table lists the mean and std. value across all datasets and executions. Baselines that did not reduce the AutoML times (i.e., their execution took longer than Full-AutoML) are omitted from the table.  

First, see that \textit{the only method obtaining a higher relative accuracy than 95\% is \system{}, with mean scores of 98.71\% for Auto-Sklearn and 97.17\% for TPOT, while reducing their original running times by 84.91\% and 76.57\%, respectively.} 

We next analyze the rest of the baselines' results w.r.t. the  categories and questions, as detailed in Section~\ref{sec:exp_baseline_desc}.

\noindent(i)  The simple Monte Carlo baseline MC-100, was naturally the fastest, yet obtained poor accuracy results ($\sim$70\%). The slower instance MC-100K, which chooses the best out of 100K random DSTs obtained slightly better accuracy ($\sim$80\%).
\textit{This shows that a trivial random sampling approach is highly ineffective, as the ML pipelines result in poor accuracy scores} (Recall that a drop of 5\% in the final model accuracy is unacceptable for most ML use-cases).

\vspace{1mm}
\noindent(ii) All baselines in Categories A-C (i.e., MC, MAB and Greedy baselines) attempt to generate entropy-preserving DSTs. 
As expected, the more sophisticated MAB surpasses the simple MC and greedy approaches with relative-accuracy scores of 92.88\% and 92.66\% for AutoSklearn and TPOT (MC and greedy scores are omitted, as they took longer than 24 hours to run).
\textit{However, \system{} significantly outperforms all these alternative optimizations, proving the efficiency of our genetic based \algo{} algorithm.}
Moreover, the only alternative optimization that achieved relative-accuracy higher than 95\% was MC-24H, which was able, after 24 hours of execution to obtain a relative-accuracy of 95.95\% (which is still lower than \system{}).

\begin{figure}[t]
    \centering
    \includegraphics[width=0.8\columnwidth]{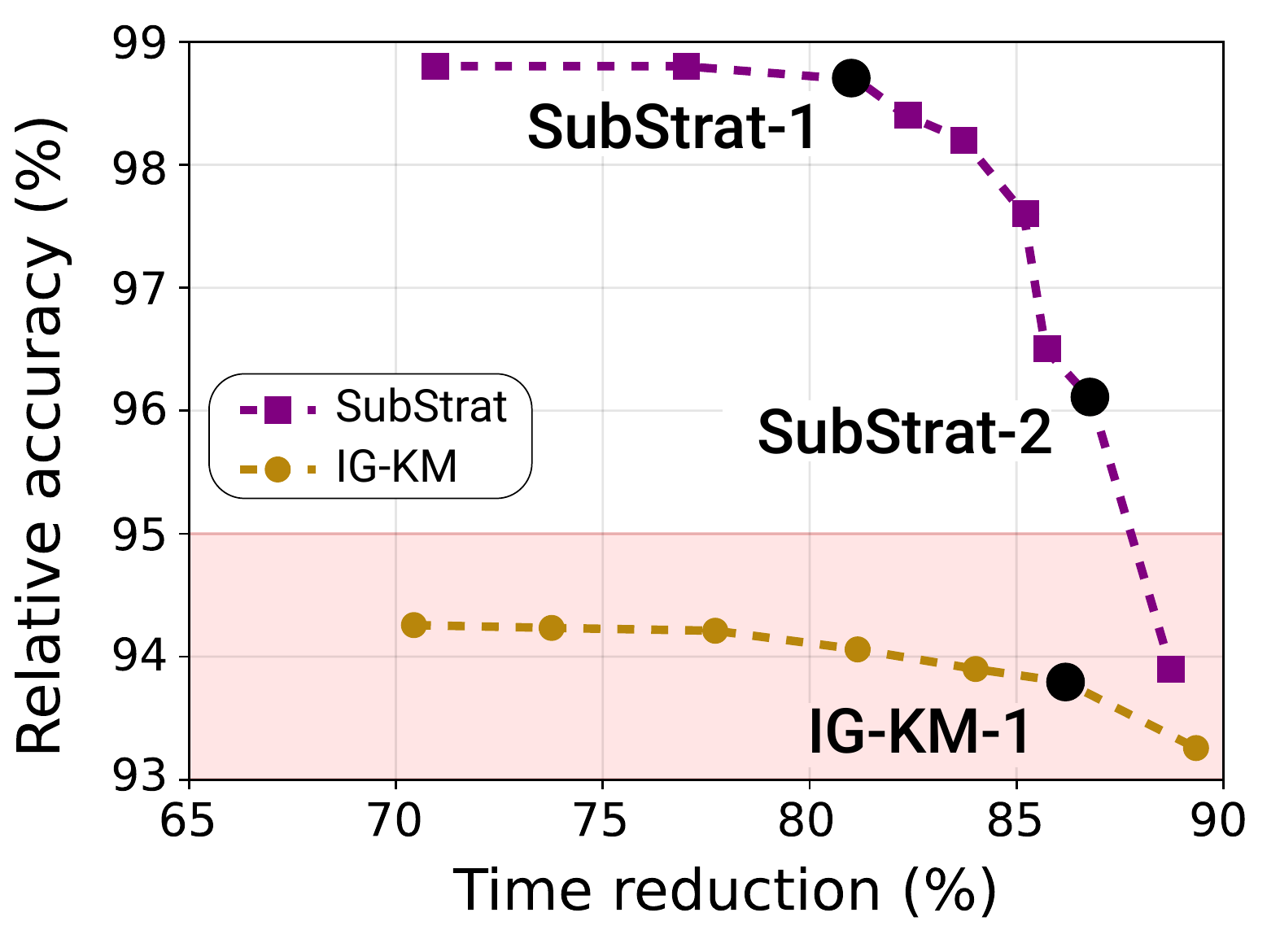}
    \vspace{-3mm}
    \caption{ {\small \system{} settings Skyline}}
    \vspace{-3mm}
    \label{fig:skyline}
\end{figure}

\vspace{1mm}
\noindent(iii) Next, we inspect the performance baselines that combine existing techniques for row sampling and column selections (Categories D and E). The best results in these categories were obtained by IG-KM, which combines information-gain feature selection and clustering-based row selection. While its time-reduction scores are slightly better (86.63\%) and (85.80\%) than \system{}, see that its relative-accuracy score are inferior (93.83\%, 94.42\%), and below the 95\% bar. \textit{These results show that generating a DST by combining row-sampling and column selection techniques is ineffective compared to \system{}}. Nevertheless, one may argue that in some cases relative accuracy slightly below 95\% is borderline acceptable. To this end, in Section~\ref{sec:exp_tradeoff} we discuss the tradeoff between time-reduction and accuracy, using different \system{} configurations. We particularly show a dominating configuration of \system{}, surpassing IG-KM in both accuracy and time-reduction (See Figure~\ref{fig:skyline} ).

\vspace{1mm}
\noindent(iv) Last, we validate the effectiveness of the fine-tuning phase.
See that \system{}-NF -- which does not perform the fine-tuning on the full dataset -- is indeed faster than \system{}, but its relative accuracy scores drop about 5\% from 98.71\% and 97.17\% to 93.09\% and 92.03\%. \textit{This proves the importance of fine-tuning the configuration on the full dataset}.
Note that similar reductions were observed for the rest of the baselines when removing the fine-tuning phase (results omitted for space constraints).

\vspace{2mm}
To shed more light on the performance of the baselines on each dataset, we illustrate the individual relative-accuracy and time-reduction scores in Figure~\ref{fig:scatter}. Each point in the plot depicts the performance of a baseline for a given dataset when using Auto-Sklearn (similar results were obtained for TPOT, thus omitted).
The light-red zone highlights all instances with less than 95\% accuracy. See that \system{}, while having some variation in time-reduction, surpasses the 95\% accuracy bar in 8 out of 10 datasets. This is significantly better than the other baselines, achieving a maximum of 3/10 datasets with more than 95\% relative accuracy (achieved by IG-KM, as illustrated in Figure~\ref{fig:scatter}). 

\begin{figure*}[t]
\vspace{-4mm}
    \centering
    \begin{subfigure}[t]{0.4\textwidth}
    \includegraphics[width=1\textwidth]{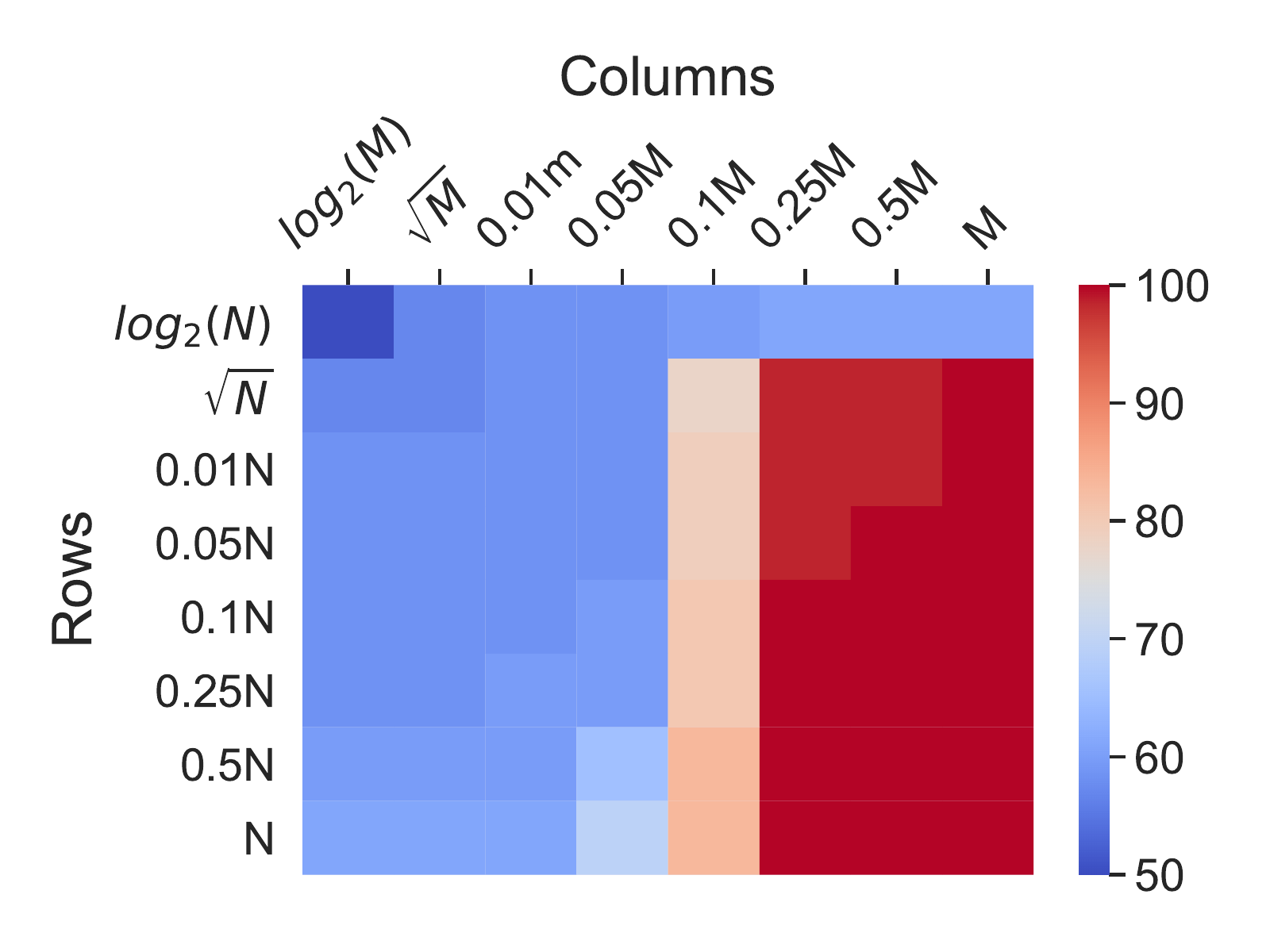}
    \vspace{-8mm}
    \caption{Relative Accuracy (\%)}
    \label{fig:heatmap_accuracy}
    \end{subfigure}
    \hspace*{9mm}
    \begin{subfigure}[t]{0.4\textwidth}
    \centering
    \includegraphics[width=1\textwidth]{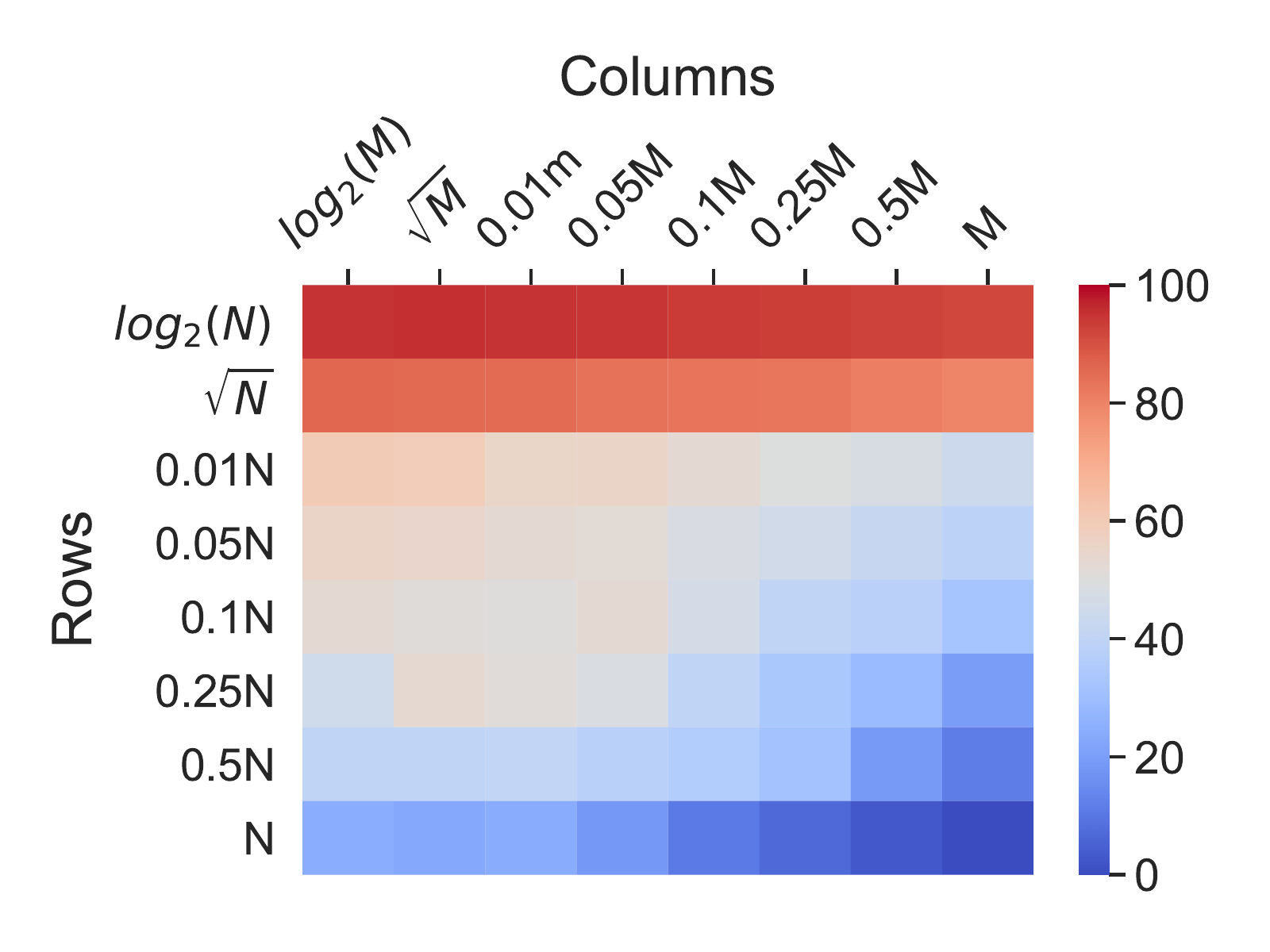}
    \vspace{-8mm}
    \caption{Time Reduction (\%)}
    \label{fig:heatmap_time}
    \end{subfigure}
    \vspace{-3mm}
    \caption{Overall Effect of DST Size}
    \label{fig:heatmaps}
\end{figure*}

\begin{figure*}[t]
    \centering
     \begin{subfigure}[t]{0.38\textwidth}
    \includegraphics[width=1\textwidth]{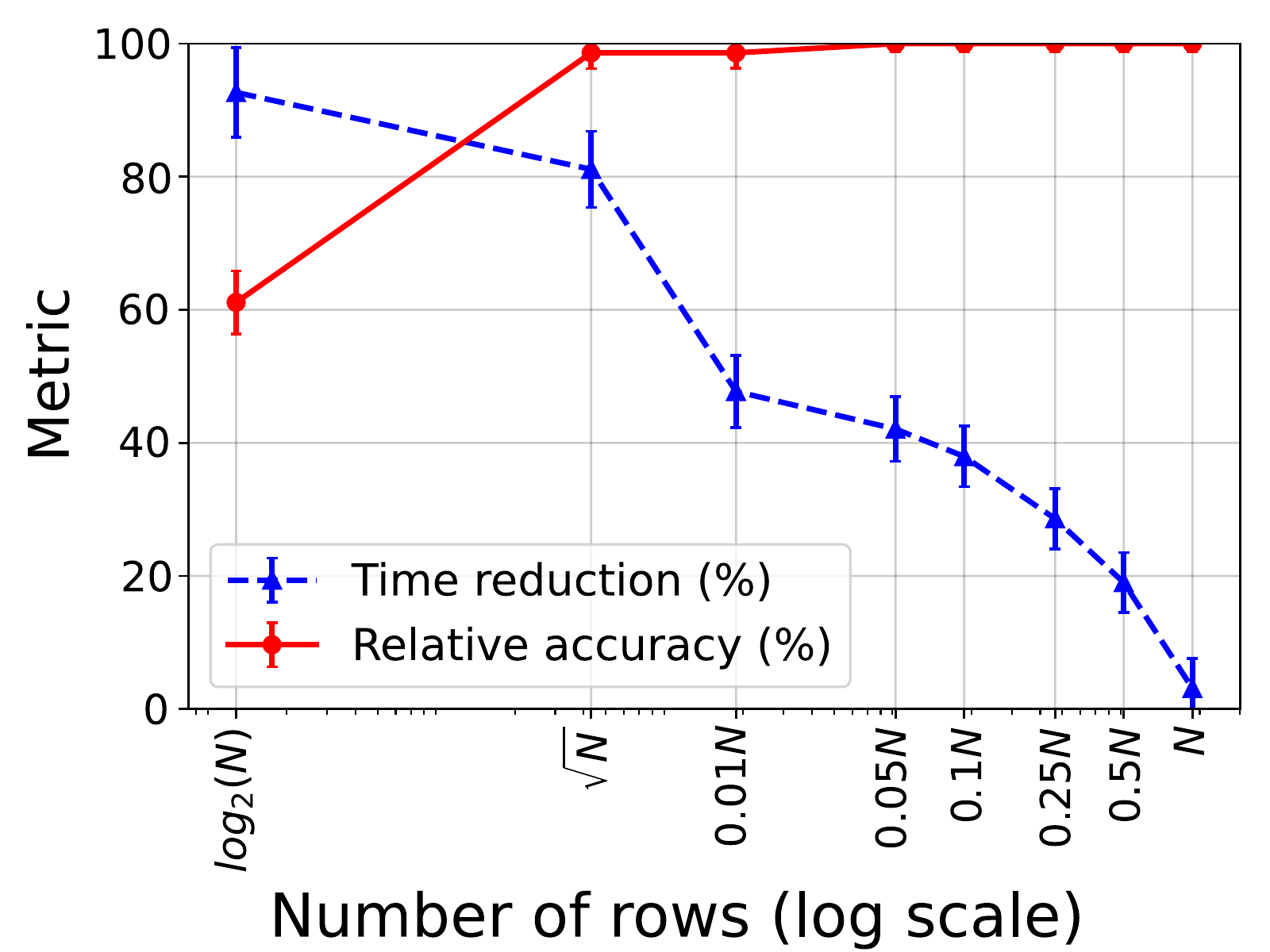}
    \caption{DST Length (\# Rows)}
    \label{fig:length_effect}
    \end{subfigure}
     \hspace*{10mm}
     \begin{subfigure}[t]{0.38\textwidth}
    \centering
    \includegraphics[width=1\textwidth]{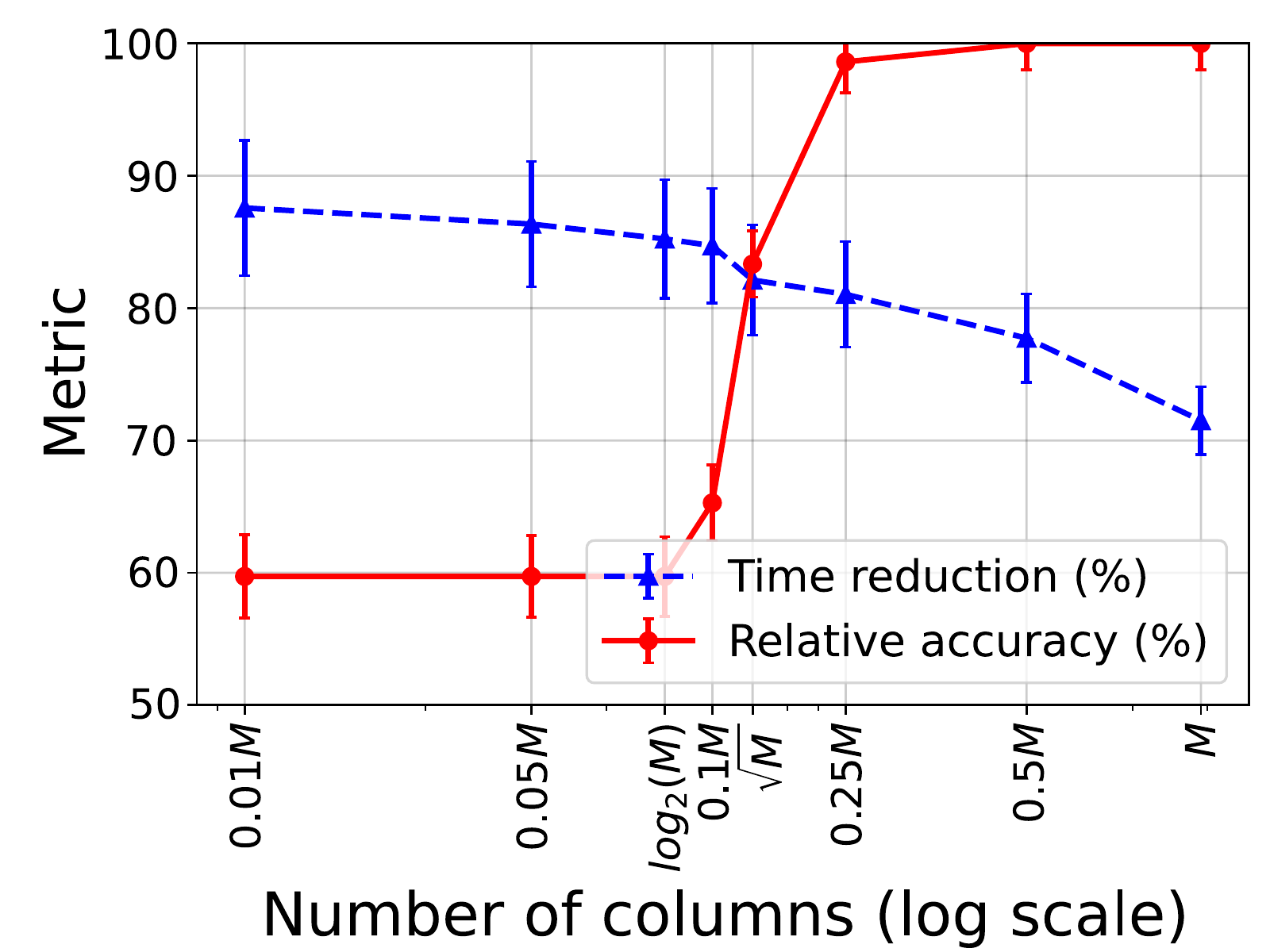}
    \vspace{-3mm}
    \caption{DST Width (\# Columns)}
    \label{fig:width_effect}
\end{subfigure}
\vspace{-2mm}
\caption{Isolated Effect of DST Length and Width}
\label{fig:isolated_effect}
\end{figure*}

\subsection{Time-Reduction \& Accuracy Trade-off}
\label{sec:exp_tradeoff}
As is also clear from the baseline comparison above, there is a natural trade-off between time-reduction and accuracy. While we assume that 95\% relative accuracy is a lower bound, users may still have slightly different needs and prefer, e.g, to further compromise on accuracy in order to obtain better speedup. 

We therefore discuss the performance of other \system{} configurations that produce higher (lower) time reductions -- at the expense of lower (higher) relative accuracy. 

Figure~\ref{fig:skyline} depicts the
performance of different configurations of \system{}, as well as for IG-KM (the best performing baseline approach) when using AutoSklearn (Similar trends were observed for TPOT, thus omitted).
For better presentation, we omitted configurations that were strictly outperformed by others (i.e., in both time-reduction \textit{and} accuracy), remaining with the performance skyline~\cite{borzsony2001skyline} of both \system{} and IG-KM.
Like in Figure~\ref{fig:scatter}, the red zone marks the area of 95\% accuracy or less.  

First, observe that \system{} provides a more flexible performance range than IG-KM: 
Our default setting labeled `\system{}-1', obtaining 98.71\% relative accuracy and time-reduction of 81.1\%,
whereas the configuration marked by `\system{}-2', which is the fastest one above the 95\% bar -- obtains a time-reduction of 86.86\%, (namely, 13X faster than Full-AutoML), while retaining 96.43\% of its accuracy.

In particular, see that `\system{}-2' strictly outperforms `IG-KM-1' (the default configuration of IG-KM), with lower performance of 86.63\% (time) and 93.83\% (accuracy). 

\subsection{Effect of DST Size (Length and Width)}
\label{sec:exp_dst_size}

Next, we examine how the choice of $n$ and $m$, i.e., the number of rows and columns in our DST, affect the performance of \system{}.  

Figure~\ref{fig:heatmaps} shows for each combination of $(n,m)$, ranging from $(log_2 N,log_2 M)$ to $(N,M)$, the average relative accuracy (Figure~\ref{fig:heatmap_accuracy}) and time reduction (Figure~\ref{fig:heatmap_time}) across all datasets when using AutoSklearn (similar trends appear for TPOT as well). In both heat-maps, the darker the red color, the better the score. 
We can see that (1) The DST width (num. of columns), needs to be at least $0.1M$ to obtain a significant improvement in accuracy, but a maximal accuracy can be obtained with DSTs with widths of $0.25$ and above. (2) The time-reduction significantly decreases when using a DST with more than $\sqrt{N}$ rows. Interestingly, this DST size also yields the best overall performance for the baselines IG-KM, MAB, IG-RAND, and MC-100 (as resulted in their hyper-parameters grid search). 

Next, Figure~\ref{fig:isolated_effect} shows the isolated effect of the length and width of the DST. Figure~\ref{fig:length_effect} shows the effect on time/accuracy performance when varying $n$, where $m$ is set to the default value of $0.25M$, and Figure~\ref{fig:width_effect} shows the effect of different $m$ values, when setting $n$ to $\sqrt{N}$. The error vertical lines depict a $.95$ confidence interval. 

We observe that increasing the DST length $n$ from $\sqrt{n}$ is clearly not worthwhile, as the time-reduction decreases significantly, while the accuracy is only marginally improved. As for increasing the DST width $m$ after $0.25M$, the trend is not as clear, but the decrease in time-reduction is visibly stronger than the accuracy improvement. 

\section{Conclusion \& Future Work}
\label{sec:conclusion}
We present \system{}, a subset-based optimization strategy for AutoML. To the best of our knowledge \system{} is the first work suggesting to intelligently reduce the data size rather than the configuration search-space in AutoML. 
This allows data scientists to keep using their favorite AutoML libraries, while significantly reducing their running times and computational costs.

Our experimental results show the inadequacy of simple solutions 
such as randomly sampling the data before applying AutoML, or employing existing techniques for row sampling and feature selection. In fact, we show that \system{}, which obtained 98\% average relative accuracy, while reducing 79\% of the computational costs, was the only approach that was able to exceed 95\% relative-accuracy. 

In future work, we intend to investigate the application of more sophisticated dataset measures, as well as additional fine-tuning strategies. Furthermore, applying meta-learning for automatically selecting the DST size as well as other \system{} parameters is also a promising direction for future research.


\end{document}